\documentclass[10pt,twocolumn,letterpaper]{article}

\usepackage{cvpr}              % To produce the CAMERA-READY version
\usepackage{graphicx}
\usepackage{amsmath}
\usepackage{amssymb}
\usepackage{booktabs}
\usepackage{arydshln}

\usepackage{times}
\usepackage{epsfig}
\usepackage{tabu}
\usepackage{multirow}
\usepackage{amsfonts}       % blackboard math symbols
\usepackage{xpunctuate}
\usepackage{pifont}% http://ctan.org/pkg/pifont
\usepackage{color,colortbl}
\usepackage[rgb,dvipsnames]{xcolor}
\usepackage{subfig}
\usepackage{tabulary,overpic}
\usepackage[normalem]{ulem}
\usepackage[accsupp]{axessibility}

\newcommand{\cmark}{\ding{51}}%
\newcommand{\xmark}{\ding{55}}%

\definecolor{fyxcolor}{RGB}{0,128,255}
\definecolor{demphcolor}{RGB}{100,100,100}
\definecolor{citecolor}{RGB}{0,0,192} % \definecolor{citecolor}{RGB}{34,139,34}
\definecolor{GrayBG}{gray}{0.95}

\def\x{$\times$}
\newcommand{\demph}[1]{\textcolor{demphcolor}{#1}}
\newcommand{\app}{\raise.17ex\hbox{$\scriptstyle\sim$}}
\newlength\savewidth\newcommand\shline{\noalign{\global\savewidth\arrayrulewidth
  \global\arrayrulewidth 1pt}\hline\noalign{\global\arrayrulewidth\savewidth}}

\usepackage[pagebackref=true,breaklinks=true,letterpaper=true,colorlinks,bookmarks=false]{hyperref}

\usepackage[capitalize]{cleveref}
\crefname{section}{Sec.}{Secs.}
\Crefname{section}{Section}{Sections}
\Crefname{table}{Table}{Tables}
\crefname{table}{Tab.}{Tabs.}

\def\cvprPaperID{9291} % *** Enter the CVPR Paper ID here
\def\confName{CVPR}
\def\confYear{2022}

\begin{document}

%%%%%%%%% TITLE
% \title{Self-supervised Pretraining for Hierarchical Multimodal Movie Understanding \fy{Q -- do we want to emphasize \emph{movie} understanding?}}
\title{Hierarchical Self-supervised Representation Learning for Movie Understanding}
% \fy{Q -- do we want to emphasize \emph{movie} understanding?}

\author{Fanyi Xiao\thanks{\it Work done while at Amazon, now at Meta AI}, $\;$ Kaustav Kundu, $\;$ Joseph Tighe, $\;$ Davide Modolo\\
AWS AI Labs\\
{\tt\small \{kaustavk,tighej,dmodolo\}@amazon.com}
}
\maketitle

%%%%%%%%% ABSTRACT
\vspace{-2mm}
\begin{abstract}

% Unlike standard video recognition models widely studied in literature, movie understanding models are different in two main aspects: 1) they require modeling capability at different levels, to solve tasks ranging from low-level ones like predicting actions in events that mostly requires short-term semantic and motion understanding, to higher level tasks like semantic role labeling which requires modeling long-term interactions across short event segments, and 2) they require modeling capability for multimodal inputs like vision and language, which are both important for comprehensive movie understanding. Therefore, we believe a movie understanding model should be hierarchical and multimodal, with different levels focusing on solving different tasks that have different modality inputs. 
% Recognizing the importance of applying the appropriate method to train different levels of the model, we propose a hybrid strategy that involves both a contrastive learning task, to pretrain low-level video features, and a event-level mask prediction task, to pretrain a feature contextualizer. We demonstrate its effectiveness across several movie understanding tasks.   
\vspace{-1mm}
Most self-supervised video representation learning approaches focus on action recognition. In contrast, in this paper we focus on self-supervised video learning for movie understanding and propose a novel hierarchical self-supervised pretraining strategy that separately pretrains each level of our hierarchical movie understanding model (based on~\cite{sadhu-vidsitu2021}).
Specifically, we propose to pretrain the low-level video backbone using a contrastive learning objective, while pretrain the higher-level video contextualizer using an event mask prediction task, which enables the usage of different data sources for pretraining different levels of the hierarchy.
We first show that our self-supervised pretraining strategies are effective and lead to improved performance on all tasks and metrics on VidSitu benchmark~\cite{sadhu-vidsitu2021} (\eg, improving on semantic role prediction from 47\% to 61\% CIDEr scores). We further demonstrate the effectiveness of our contextualized event features on LVU tasks~\cite{wu-cvpr2021}, both when used alone and when combined with instance features, showing their complementarity.

% and how to combine it with instance features to improve on LVU benchmark tasks~\cite{wu-cvpr2021}.
% and further demonstrate the effectiveness of our contextualized event features and how to combine it with instance features to improve on LVU benchmark tasks~\cite{wu-cvpr2021}.
% we conduct an empirical study to understand how effective different self-supervised video pretraining methods are, for different levels of a hierarchical model. 

\end{abstract}

%%%%%%%%% BODY TEXT
\vspace{-4mm}
\section{Introduction} \label{sec:intro}

% \dvd{we need to talk about the intro. the flow doesn't work. we focus a lot of attention of what's important for movies (i.e., hierarchical modelling), but we can't do that since [34] already did that. instead, we should really mostly focus our attention on the importance of pretraining. maybe we can discuss the importance of pretraining on classic action tasks and how pretraining is even more critical for movies that require high-level contextualization. }

% For example, contrastive learning has emerged as a particularly effective framework to encourage the learning of intra-instnace invariances~\cite{}, which is criticial for action recognition. 

Most of the latest research on self-supervised video representation learning (SSL) focuses on the task of action recognition~\cite{diba-dynamonet2019,benaim-speednet2020,qian-cvrl2020,han-coclr20,xiao-modist2021,feichtenhofer-cvpr2021,recasens-brave2021}. This priority has largely influenced the design of these methods, as well as the type of datasets used to learn their representations.  
For example, they propose models that encourage the learning of short-term appearance and motion cues, as these are the most informative for action recognition. At the same time, they mostly focus on pretraining on the Kinetics~\cite{kay-kinetics2017} dataset, which consists of hundreds of thousands of short YouTube clips with diverse motion and semantic patterns.  
Unlike these works, we are interested in learning self-supervised video representations to understand movies.

\begin{figure}[t]
    % \centering
    \hspace{-5mm}\includegraphics[width=0.55\textwidth]{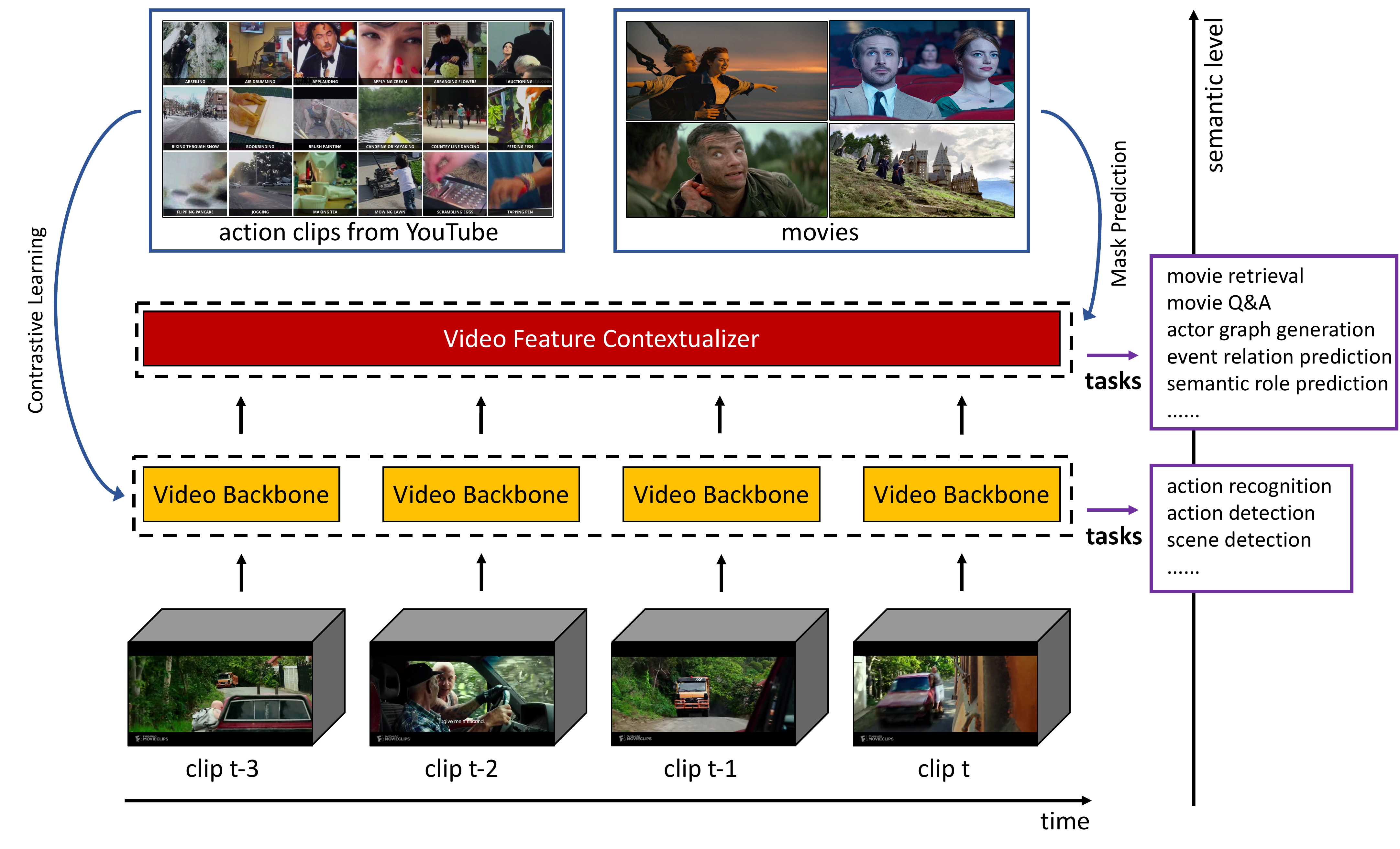} 
    \vspace{-5mm}
    \caption{ \small \it 
    \textbf{Hierarchical self-supervised pretraining}. 
    We pretrain the low-level video feature backbone using contrastive learning objectives on large collections of YouTube-style action clips; while pretrain the higher-level feature contextualizer using mask prediction on movies with rich temporal plots.  
    % \fy{Add the text label specifying the pretraining task}
    }
    \label{fig:concept}
    \vspace{-5mm}
\end{figure}

Movies are however very complex and they require reasoning at many levels: from the simple understanding of low-level actions to the interpretation of high-level semantic narratives, which require knowledge of the characters, their histories, relationships, behaviours, etc. Towards building rich models for movie understanding, \cite{sadhu-vidsitu2021} recently proposed a hierarchical movie understanding model that learns in a fully supervised fashion. However, it is extremely difficult to annotate large-scale video datasets, even for a relatively simple task like action classification, not to mention for complex movie tasks (\eg labeling actor relation graphs~\cite{vicol-moviegraphs2018}). % Annotating large-scale video datasets is extremely difficult for a relatively simple task like action recognition. It is even more challenging for complex movie tasks (\eg labeling actor relation graphs~\cite{vicol-moviegraphs2018}).
To overcome this bottleneck, we propose a novel hierarchical self-supervised pretraining strategy that separately pretrains each level of this hierarchical model.

In details, the hierarchical movie model of ~\cite{sadhu-vidsitu2021} consists of two levels: a low-level video backbone encoder and a higher-level transformer contextualizer (Fig.~\ref{fig:concept}). We design our hierarchical learning strategy to sequentially pretrain the backbone and the transformer encoder as they specialize in different aspects of movie understanding.
The backbone is responsible for the heavy-lifting work to extract low-level appearance and motion cues for people, objects and scenes from raw pixels. Therefore, it needs to be high in capacity and can be trained on a large amount of YouTube videos (e.g., Kinetics~\cite{kinetics}). Once we obtain an appropriate feature abstraction from the video backbone, we can treat such representations as visual ``word tokens'' and learn to contextualize the neighboring visual tokens. The contextualizer can be lightweight and trained on a small amount of training data with stronger semantic and temporal structures (i.e., movies). 
% Ideally, all the levels could be trained on videos belonging to the target domain. However, the data in the movie domain is limited and often insufficient. 

% \sout{As the video backbone needs to do the heavy-lifting of extracting low-level appearance and motion cues for people, objects and scenes, we propose to pretrain it using a \emph{contrastive learning} objective, which is a very effective pretraining paradigm for action recognition~\cite{qian-cvrl2020,han-coclr20,xiao-modist2021,feichtenhofer-cvpr2021,recasens-brave2021}, as it is tailored to help models learn the intra-instance invariances from visual cues.
% On the other hand, once we obtain an appropriate feature abstraction from the video backbone, we can treat that like visual ``word tokens'' and pretrain the higher level transformer model to produce contextualized semantic representations using the \emph{mask prediction} task, which has been shown effective for pretraining language models that take in word tokens for contextualization~\cite{devlin-bert2018,liu-arxiv2019}.}

In details, we propose to pretrain the video backbone using a \emph{contrastive learning} objective, which helps models learning the intra-instance invariances from visual cues. This pretraining paradigm has shown to be very effective for action recognition~\cite{qian-cvrl2020,han-coclr20,xiao-modist2021,feichtenhofer-cvpr2021,recasens-brave2021}. Furthermore, we pretrain the higher level transformer model to produce contextualized semantic representations using the \emph{mask prediction} task, which has been shown effective for pretraining language models that take in word tokens for contextualization~\cite{devlin-bert2018,liu-arxiv2019}. These hierarchical self-supervised pretraining strategies bring two data advantages: they enable the use of different data sources for pretraining the different levels of the hierarchy, and they do not require any annotation, which are inherently expensive to collect.

We evaluate the impact of our pretrainings on the recently released VidSitu~\cite{sadhu-vidsitu2021} and LVU~\cite{wu-cvpr2021} datasets. These are movie datasets that have been annotated for various tasks, ranging from low-level verb prediction (i.e., actions) to high-level semantic role prediction or event relation classification (i.e., ``A causes B''). Our results show that our self-supervised pretraining strategies are effective and lead to improved performance on all tasks and metrics. For example, on the task of Semantic Role Prediction, we improve CIDEr~\cite{vedantam-cider2015} metric performance over the previous, fully-supervised, state-of-the-art~\cite{sadhu-vidsitu2021} from 47\% to 61\%. Finally, we also ablate the design choices of our pretraining recipes.

\section{Related Work} \label{sec:relatedwork}

\paragraph{Self-supervised video representation learning.} 
Many works have explored ways to learn representations by designing pretext tasks that exploit the temporal structure of videos. For example, some works attempted to learn representations by predicting the ordering of video frames~\cite{misra-eccv2016,fernando-cvpr2017}, while others instead designed the task of predicting direction~\cite{wei-cvpr2018} and speed~\cite{benaim-speednet2020} of the video.  Others attempted to learn video representations by either tracking across frames patches~\cite{wang-iccv2015}, pixels~\cite{wang-cvpr2019}, colors~\cite{vondrick-color2018}, or by predicting temporal context for videos~\cite{diba-dynamonet2019,recasens-brave2021,wang-lstcl2021}. A more recent line of work overcomes the need for pretext tasks by leveraging the \emph{contrastive learning} paradigm~\cite{qian-cvrl2020,han-coclr20,feichtenhofer-cvpr2021,xiao-modist2021} and achieved impressive results even when comparing to fully-supervised methods. Though flourishing, all of the works mentioned above focus on learning video representations from short YouTube-style action clips (\eg, Kinetics) and have action recognition as the task in mind when designing learning objectives and architectures. In contrast, we are interested in learning video representations \emph{from movies} and \emph{for movies}, which as we will elaborate in next sections, require very different learning objectives and architectures. 

From this perspective,~\cite{sun-videobert2019,wu-cvpr2021,tan-vimpac2021} are closest to our work in that they also pretrain a transformer for feature contextualization. However,~\cite{wu-cvpr2021,tan-vimpac2021} focus on masking spatial regions, either the object boxes in~\cite{wu-cvpr2021} or small patches in~\cite{tan-vimpac2021}, to learn the spatial arrangements in videos. Whereas~\cite{sun-videobert2019} relies on joint video and language masking, which requires aligned video-narration pairs. 
% which requires preprocessing using object detectors trained with box supervision, whereas~\cite{sun-videobert2019} form input visual tokens through a vector quantization step using hierarchical k-means clustering. 
In contrast, we demonstrate we can directly pretrain our contextualizer with mask prediction using a simple event-level representation, without needing any object detectors trained with box supervision, or video-text pairs. 

% \kaustav{Compare shot level masking w.r.t. object level masking~\cite{wu-cvpr2021}, frame level masking (VideoBERT).}

\begin{figure*}[t]
    \centering
    % \hspace{-5mm}
    \includegraphics[width=1.0\textwidth]{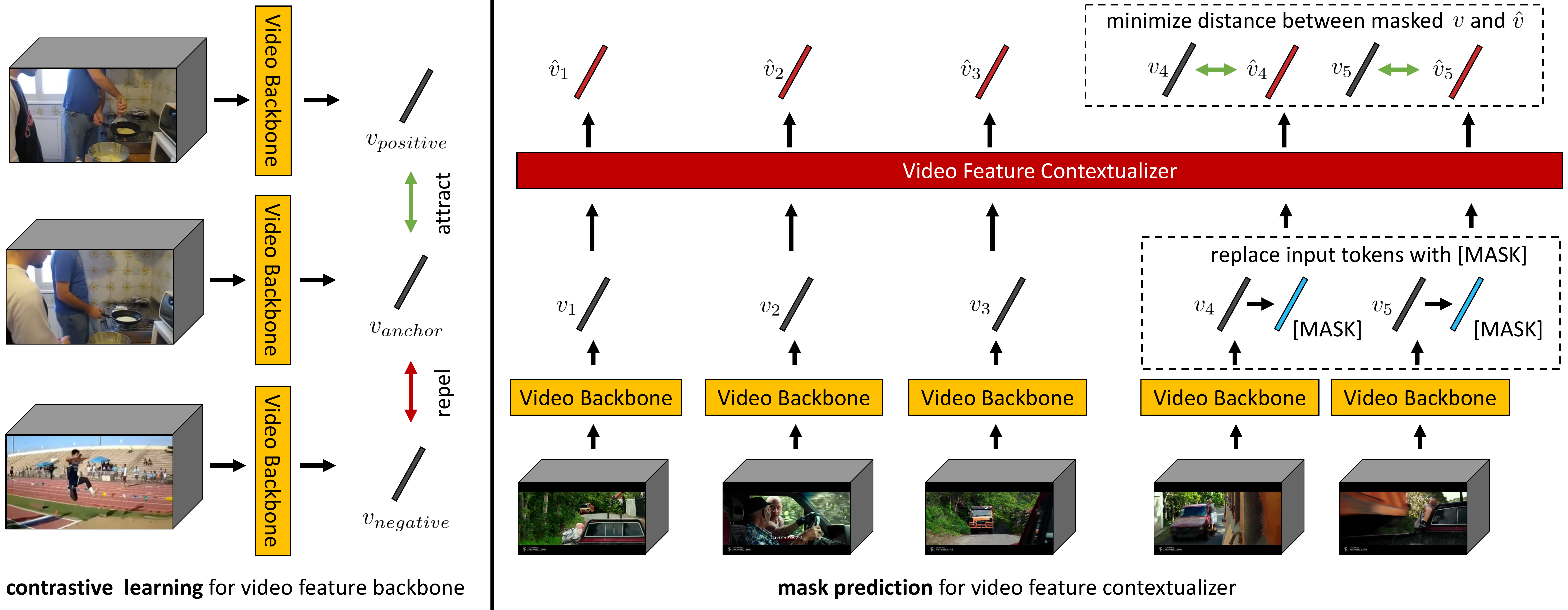} 
    \caption{\small  \it 
    \textbf{Overview of our hierarchical pretraining methods}. 
    The \textbf{left} shows how we use contrastive learning to pretrain our video feature backbone --- features $v_{anchor}$ and $v_{positive}$ produced from two clips of the same video are pulled together to each other, whereas the feature $v_{negative}$, computed from a clip sampled from another video, is pushed away.
    Whereas the \textbf{right} shows how we pretrain our feature contextualizer using mask prediction --- in this sequence of 5 tokens, we mask out input tokens $v_2$ and $v_3$ to the contextualizer, and then forward through to get the outputs $\hat{v}_i$. We then set the learning objective to minimize the distance between the output tokens ($\hat{v}_2$, $\hat{v}_3$) and the masked-out input tokens ($v_2$, $v_3$).   
    % http://motion.aka.corp.amazon.com:8200/workplace/data/motion_efs/datasets/VidSitu/vsitu_video_trimmed_dir/v_-0SHIbuEO3w_seg_55_65.mp4
    % \joe{i'm not a big fan of having '/' symbolize a token. is this common in language? also you use the same video as in modist, you might want to change it so it doesn't point to the same set of authors.}
    }
    \label{fig:approach}
    \vspace{-5mm}
\end{figure*}

\vspace{-5mm}
\paragraph{Movie understanding.} 
Researchers have explored many individual movie understanding tasks, including low-level tasks like spatio-temporal action detection~\cite{gu-ava2018}, scene detection~\cite{huang-movienet2020}, metadata classification (\eg, genre)~\cite{wu-cvpr2021}, as well as tasks that require higher-level contextualization and reasoning like movie description~\cite{rohrbach-dataset2015}, movie question answering~\cite{tapaswi-movieqa2016}, story-based retrieval~\cite{bain-condensed2020}, semantic role prediction~\cite{sadhu-vidsitu2021} and social graph generation~\cite{vicol-moviegraphs2018}. 
% This includes tasks like movie content retrieval (CondensedMovies),  Tapaswi et al. movie question answering. \dvd{also add movie classification, scene detection, etc.?} 
Unlike these work which mostly focus on a single task, we demonstrate the general benefits of our pretraining strategy by transferring it to a hierarchy of movie tasks. 

\vspace{-5mm}
\paragraph{Contextualized temporal modeling for videos.} One distinct characteristic of movie understanding is that there exists strong semantic correlation across neighboring scenes and events~\cite{na2017read, tapaswi-movieqa2016}. 
% \sout{Several works try to model this in different ways.}
An effective way to learn temporal contextualization is to apply RNNs to model the evolution of frames~\cite{donahue-cvpr2015,ng-cvpr2015,sun-iccv2017,li-recurrent2018,li-videolstm2018}. To deal with longer temporal window, it's helpful to establish an explicit feature bank to store useful features along time~\cite{wu-lfb2019}. 
To model finer-grained interactions, there are works that leverage pre-computed object/person proposals or detections~\cite{wang-eccv2018,baradel-eccv2018,ma-attend2018,sun-eccv2018}. Though related, our focus is different in that we're interested in developing effective pretraining methods for feature contextualization, and also we want to achieve so without using external object or person detectors for preprocessing.

\section{Hierarchical SSL for Movies} \label{sec:approach}
% intro
Understanding movies is a complex task and it requires reasoning at many levels. Towards learning rich representations for movies, \cite{sadhu-vidsitu2021} was the first work to propose a hierarchical model that consists of a low-level CNN video feature backbone and a high-level transformer feature contextualizer. Inspired by it, this work aims at delving deeper into hierarchical movie understanding and explore the importance of pretraining for movies.
Specifically, we re-evaluate the choices of \cite{sadhu-vidsitu2021} (Sec.~\ref{sec:hierarchical_model}), which pretrained their video backbone on a fully supervised action dataset and trained the contextualizer from scratch, and demonstrate that our pretraining strategies are better designed to help each level of the hierarchy learn features that are meaningful for the task of movie understanding. Since each level is responsible for different goals, we propose to separate the pretraining of the video backbone and the feature contextualizer.

% encoder
As the video backbone is responsible for extracting low-level appearance and motion cues, we propose to pretrain it using self-supervised contrastive learning (Sec.~\ref{sec:video_backbone}), which explicitly models the intra-instance invariances. 
%
% contextualizer
On the other hand, the video feature contextualizer is responsible for propagating information across neighboring clips (``visual tokens''). Inspired by the NLP literature~\cite{devlin-bert2019}, we propose to pretrain it using an event-level mask prediction task (Sec.~\ref{sec:contextualizer}), with some key differences to how it's applied in previous approaches.
In contrast to some recent works in computer vision that propose to apply mask predictions to learn \emph{spatial arrangements} of video patches or objects~\cite{tan-vimpac2021,wu-cvpr2021}, we focus on learning the \emph{temporal contextualization} for event representations. Furthermore, unlike~\cite{sun-videobert2019}, which requires joint video and language masking, we show that our method learns strong contextualized event features using only video clips. Since our method does not need any object detector~\cite{wu-cvpr2021} or synced video-narration pairs~\cite{sun-videobert2019}, it makes our method simpler and more scalable.

Finally, this decouping also enables us to pretrain the different levels on different datasets. This leads to better specialization (since we can use the most suitable dataset for each level) and less reliance on having large-scale datasets for the target domain (i.e., we do not require hundreds of thousands of movies to train the expensive video backbone). This is an important advantage over previous SSL methods that mostly conduct pretraining of the full model with a single task and dataset.

\subsection{Hierarchical model for movie understanding} \label{sec:hierarchical_model}
We follow~\cite{sadhu-vidsitu2021} and adopt their hierarchical architecture for movie understanding (Fig.~\ref{fig:concept}). It uses a 3D CNN as the low-level video feature backbone and a transformer encoder and decoder for feature contextualization and natural language generation, respectively.
The video backbone extracts features $v_t$ on short 2-seconds clips. Then, the transformer encoder operates on a sequence $\{v_t, v_{t+1},...,v_N\}$ of consecutive clip features and contextualizes them into $\hat{v}_t$. These contextualized features are finally used as input to either a classifier (\eg for visual tasks, like event relation prediction), or a transformer decoder (TxD), that decodes natural language outputs (\eg, for semantic role prediction). 

For the backbone, we adopt the popular Slow-only network~\cite{slowfast}, but with two modifications \cite{qian-cvrl2020}: 1) instead of 8\x8 inputs (8 frame inputs sampled 8 frames apart), we use denser 16\x4 inputs with a temporal kernel of 5 for the first conv layer to increase its temporal receptive field; 2) the inputs are downsampled with a temporal stride of 2 after the first conv layer, to save computation. We denote this backbone as Slow-D for its denser inputs.

For the transformer encoder (TxE) and decoder architecture (TxD), we largely follow~\cite{sadhu-vidsitu2021}. Specifically, for the transformer encoder, we use 3 layers of multi-head self-attention with residual connections, with 16 heads each and a hidden dimension of 1024. For inputs, we append a learned position embedding to each of $N$ tokens in the input sequence, which we found to work better compared to the sinusoidal embedding used in~\cite{sadhu-vidsitu2021}. The transformer decoder also has 3 layers,  each one consisting of a self-attention module and a cross-attention module, where the self-attention is computed for text inputs only, while the cross-attention is added for the text tokens to query visual tokens as keys. 
For more details on the architecture, please refer to our supp. materials. %\joe{this is a lot of detail upfront, should we move some to the implementation details?} \fy{when thinking more about this, I feel it's actually not good to put this into implementation details as that's under VidSitu experiments. Maybe it's okay to leave it here? } \dvd{yeah, I see Joe's point, but considering that we don't have an explicit impl details section outside of vidsitu, I am fine with keeping it here. }

\subsection{Video backbone: contrastive pretraining} \label{sec:video_backbone}
We adopt instance discrimination contrastive learning, as it has been demonstrated to be very effective to learn visual semantics patterns by capturing the intra-instance invariances~\cite{chen-simclr2020,he-moco2020,qian-cvrl2020,xiao-modist2021,recasens-brave2021}. Among these, we experiment with two simple, yet robust, methods: CVRL~\cite{qian-cvrl2020} and MoDist~\cite{xiao-modist2021}. 
CVRL pretrains using the popular InfoNCE objective~\cite{oord-arxiv2018}. Their goal is to pull together the representations of two clips sampled from the same video, while pushing apart those of clips that are sampled from different videos (Fig.~\ref{fig:approach} left). 
Though it yields impressive results, CVRL does not explicitly utilize motion cues for representation learning. MoDist~\cite{xiao-modist2021} addresses this with a visual-motion cross-modal contrastive objective where a supporting motion network is used to distill information to the visual backbone, so that it can learn motion-sensitive features. 

\subsection{Contextualizer: mask prediction pretraining}  \label{sec:contextualizer}
% \dvd{we probably want a figure for this} 
The goal of pretraining the transformer is to make it better at contextualizing the individual semantic tokens (video clips in our settings). 
For this, we use the mask prediction task, which is widely used in language model training in NLP (\eg, BERT~\cite{devlin-bert2019}). Specifically, as shown in Fig.~\ref{fig:approach} (right), given a set of visual tokens $\{v_1,v_2,...,v_N\}$, we randomly select a mask size $m$ from the set $m\in\{1, 2, ..., \alpha N\}$, where $\alpha \in [0, 1]$ determines the ratio of the max mask size with respect to the sequence length, and a mask starting position $s \in \{1, 2, ..., N-m+1\}$. With the sampled size and location, we mask out the selected tokens $\{v_s, ..., v_{s+m-1}\}$ and replace them with a special {\tt [MASK]} token. Then, we forward the masked sequence through the transformer encoder to obtain its $L_2$ normalized outputs $\{\hat{v}_1,\hat{v}_2,...,\hat{v}_N\}$. 
Ideally with proper contextualization from neighboring clips, even with the input clip masked out, $\hat{v}_t$ should still be able to ``fill in'' the semantic information carried in $v_t$ (\eg, predicting the ``car passing by'' scene for the last two input tokens in Fig.~\ref{fig:approach} right, given the first three tokens as contexts). Solving this problem is key to learning the rich temporal dynamic of movies and we formulate its learning objective in a way that pushes the output $\hat{v}_t$ to be close to its corresponding input $v_t$: % t \in \{s, ..., s+m-1\}
\begin{align}
\mathcal{L}_{mp} = -\log \frac{\exp(\hat{v}_t{\cdot}v_t / \tau)}{\exp(\hat{v}_t{\cdot}v_t / \tau) + \sum_{i=1}^{K}\exp(\hat{v}_t{\cdot}p_i  / \tau)}, 
\label{eq:lmp}
\end{align} 
where $\tau$ is a temperature parameter, and $p_i$ are a set of distractions, from which we would like $\hat{v}_t$ to identify $v_t$ by predicting its semantics from contexts. We construct the pool of distractions $\{p_1, p_2, ..., p_K\}$ by maintaining a FIFO queue during training. Note that a simpler alternative is to directly enforce an $L_2$ loss between $\hat{v}_t$ and $v_t$, but empirically we found this to yield slightly worse results than Eq.~\ref{eq:lmp} (Table~\ref{table:vidsituablatemaskpred}). 
Finally, note that though possible, we do not study the pretraining of the transformer decoder in this work,
% \sout{, as it involves both language and vision modalities.} 
as we are mainly interested in pretraining movie representations that can be transferred generically (i.e., backbone+TxE), whereas TxD is task-dependent (e.g., the decoder for SRL is a multimodal transformer that takes in both texts and videos) and is only used for certain tasks.
\section{Experiments: VidSitu Benchmark} \label{sec:vidsitu}

\begin{table*}[h]
  \centering
  \small
  \resizebox{2.1\columnwidth}{!}{
  \begin{tabular}{l | c c | c c c c c}
    % \shline
    \multicolumn{1}{c|}{} & \multicolumn{2}{c|}{pretraining} & 
    \\
    \multicolumn{1}{c|}{Model} & backbone &  TxE & CIDEr~\cite{vedantam-cider2015} & CIDEr-verb & CIDEr-arg & ROUGE-L~\cite{lin-rouge2004} & LEA~\cite{moosavi-acl2016}
    \\
    \shline
    % \demph{GPT2~\cite{radford-gpt2019}} & \demph{-} & \demph{-} & \hspace{0.3mm} \demph{34.67} & \hspace{0.3mm} \demph{42.97} & \hspace{0.3mm} \demph{34.45} & \hspace{0.3mm} \demph{40.08} & \hspace{0.3mm} \demph{48.08} 
    \demph{GPT2~\cite{radford-gpt2019}} & \demph{-} & \demph{-} & \demph{34.67} & \demph{42.97} & \demph{34.45} & \demph{40.08} & \demph{48.08} 
    \\
    \demph{I3D+TxD~\cite{sadhu-vidsitu2021}} & \demph{Sup-K400} & \demph{-} &  \demph{47.14} &  \demph{51.61} &  \demph{41.29} &  \demph{40.67} &  \demph{37.89} 
    \\
    \demph{I3D+TxE+TxD~\cite{sadhu-vidsitu2021}} & \demph{Sup-K400} & \demph{N} &  \demph{47.06} &  \demph{51.67} &  \demph{42.76} &  \demph{42.41} &  \demph{48.92} 
    \\
    \hline
    Slow-D+TxD & Sup-K400 & - & 51.37 ± 1.06 & 59.68 ± 0.88 & 46.10 ± 0.90 & 41.37 ± 0.59 & 36.03 ± 0.70 % vbarg_denseslow_txd_PretrainBackbone_K400Supervised_PretrainTXE_Scratch
    \\
    Slow-D+TxE+TxD & Sup-K400 & N &  51.36 ± 1.04 &  59.72 ± 0.87 &  47.25 ± 0.94 &  41.72 ± 0.65 &  45.99 ± 0.56 % vbarg_denseslow_txe_txd_feat_supervised_multiruns
    \\
    Slow-D+TxE+TxD & CVRL-K400 & N &  54.40 ± 0.96 &  63.18 ± 1.27 &  47.63 ± 1.83 &  41.80 ± 1.01 &  46.31 ± 1.13 % vbarg_denseslow_txe_txd_feat_K400_SLOW_8x8_R50_InfoNCE_train216k_epoch400_multiruns
    \\ 
    \hline
    Slow-D+TxE+TxD & CVRL-K400 & MaskPred-K400 &  57.48 ± 1.74 &  65.08 ± 1.92 &  51.21 ± 1.87 &  41.65 ± 1.15 &  45.71 ± 0.87 % vbarg_denseslow_txe_txd_PretrainBackbone_K400_SLOW_8x8_R50_InfoNCE_train216k_epoch400_PretrainTXE_K400_DENSESLOW_8x8_R50_InfoNCE_AGGTX_MaskPred_MaxEpoch40
    \\
    Slow-D+TxE+TxD & CVRL-VS & MaskPred-VS &  44.32 ± 0.56 &  52.07 ± 0.81 &  39.08 ± 0.64 &  40.56 ± 0.34 &  \textbf{48.87 ± 0.62} % vbarg_denseslow_txe_txd_PretrainBackbone_VidSitu_InfoNCE_Scratch_PretrainTXE_VidSitu_MaskPred
    \\  
    % \hline
    Slow-D+TxE+TxD & CVRL-K400 & MaskPred-VS &  60.34 ± 0.75 &  69.12 ± 1.43 &  53.87 ± 0.97 &  \textbf{43.77 ± 0.38} &  46.77 ± 0.61 % vbarg_denseslow_txe_txd_PretrainBackbone_K400_SLOW_8x8_R50_InfoNCE_train216k_epoch400_PretrainTXE_VIDSITU_DENSESLOW_8x8_R50_InfoNCE_AGGTX_MaskPred_Backbone_K400_SLOW_8x8_R50_InfoNCE_train216k_epoch400
    \\
    Slow-D+TxE+TxD & CVRL-K400 & MaskPred-LVU &  \textbf{61.18 ± 1.48} &  \textbf{69.15 ± 1.57} &  \textbf{54.99 ± 1.12} &  43.38 ± 0.87 &  47.81 ± 0.90 % vbarg_denseslow_txe_txd_PretrainBackbone_K400_SLOW_8x8_R50_InfoNCE_train216k_epoch400_PretrainTXE_LVU_DENSESLOW_8x8_R50_InfoNCE_AGGTX_MaskPred
    \\
    \hline
    \demph{Human (up. bound)} & & &  \demph{84.85} & \demph{91.70}  & \demph{80.15}  & \demph{39.77}  &  \demph{72.10}
  \end{tabular}}
  \vspace{-4mm}\caption{\small \it \textbf{Semantic role prediction results on VidSitu.}
  Results in the top section are taken from~\cite{sadhu-vidsitu2021}.
  The bottom row shows the human performance by measuring the agreement between annotators~\cite{sadhu-vidsitu2021}, which serves as the performance upper bound. 
  Unlike~\cite{sadhu-vidsitu2021}, which only reports results for a single run, we found there is large variance across runs (likely due to the fact that this task evaluates free-form natural language outputs), therefore we run 10 times for each experiment and report its mean and standard error. 
  }
  \label{table:vidsitusrl}
  \vspace{-5mm}
\end{table*}

% (\ie, action recognition for short clips)
VidSitu~\cite{sadhu-vidsitu2021} is a comprehensive movie understanding benchmark that features different tasks ranging from the low-level, visual-only ``verb prediction'' on short clips, to the higher-level, multimodal ``semantic role prediction'' and ``event relation'' classification. The dataset contains 29k 10-seconds clips from 3k different movies from MovieClips~\cite{movieclips}. The dataset provides detailed annotations for each clip, including 1) verb class labels at 2-sec intervals (i.e., each 10-sec clip is split into 5 ``events''), 2) semantic role labels for each annotated verb, and 3) labels specifying relations between two events (\eg, event A {\tt \small is caused by} event B). The dataset is split into a train set of 23.5k clips and a val set of 1.3k clips, on which we evaluate our models. 
Finally, to avoid data contamination, we remove 241 videos from VidSitu val that are overlapping with LVU dataset.  

\vspace{-2mm}
\paragraph{Implementation details.} For self-supervised pretraining of our video backbone, we use the training sets of the large-scale Kinetics-400~\cite{kay-kinetics2017} (K400, 240k clips) and the much smaller scale, but in-domain, VidSitu (23.5k). We pretrain both CVRL and MoDist for 400 epochs on K400 and, when specified, an additional 200 epochs on VidSitu.  
% This helps minimize the gap between YouTube \fy{we can emphasize from a positive perspective which is to combine best of both worlds}, for which we have a massive amount of video clips (240k), and the movie domain (VidSitu), for which we only have a small number.
We pretrain our transformer encoder only on movie clips from VidSitu and LVU~\cite{wu-cvpr2021} (which is another dataset of 10k movie clips), 
as we want the transformer encoder to learn about temporal contextualization from movies. 
We pretrain TxE with mask prediction for 100 epochs on VidSitu and 1000 epochs on LVU as the movie clips are \app10x longer than VidSitu.  
Note that we never use any human labels from the datasets for either of these pretraining (backbone and TxE). 
For contextualizer hyperparameters, we set input sequence length as $N=5$, $\alpha=0.6$, $\tau=0.1$, $K=65536$.
Please refer to our supp. materials for more details about pretraining. 

% \fy{Adding that we are working with input sequence length $N=5$, $\alpha=0.6$, $\tau=0.1$, $K=65536$}

% \noindent {\it Backbone.} We experiment with two state-of-the-art contrastive learning methods for self-supervised learning of our video backbone: CVRL~\cite{qian-cvrl2020} and MoDist~\cite{xiao-modist2021}. CVRL extends the popular image-based SimCLR~\cite{chen-simclr2020} to the video RGB-Time domain, whereas MoDist is a cross modality (RGB-Time and optical flow edges) contrastive learning framework that trains motion-sensitive video features.  

% \noindent {\it Model.} Instead of using the classic Slow-only R50 network~\cite{slowfast}, we follow \cite{qian-cvrl2020} and use dense 16\x4 inputs (rather than 8\x8) and a stride 2 convolution at the beginning of the network. We denote this as {\tt Slow-D}. 

% \noindent {\it Results.} \dvd{maybe we could explain shortly why we report variance in our tables}

\subsection{Semantic role prediction}
% (the benchmark adopt the argument format described in PropBank~\cite{})
% We have examined the effectiveness of contrastive learning to pretrain the video backbone so far. 
In this section, we study the effectiveness of self-supervised pretraining for semantic role prediction, which is a very challenging movie understanding task, due to its rich output space (free form natural language) and its multimodal nature (visual and language). The goal of this task is to predict various semantic role labels for each verb, including for example the agent and the patient of the verb, as well as other attributes like the scene where the verb is happening, and the description about how it's carried out (e.g. ``urgent''). Due to high human disagreement on some of the annotated roles, the benchmark only evaluates the agent (e.g. ``person''), the patient (e.g. ``ball''), the instrument/benefactive/attribute (e.g. ``towards a basket'') and the location/scene (see Sec. 4.1 of~\cite{sadhu-vidsitu2021} for more details). Following~\cite{sadhu-vidsitu2021}, we evaluate using the CIDEr~\cite{vedantam-cider2015} score metric (and its variants CIDEr-verb and CIDEr-arg that are CIDEr scores averaged across verbs and argument types). Furthermore, for completeness we also report ROUGE-L~\cite{lin-rouge2004} and LEA~\cite{moosavi-acl2016}. Finally, since we observe large variance across runs for this task, we run 10 times for each experiment and report its mean and standard error.

\begin{table*}[h]
  \centering
  \small
  \begin{tabular}{l | c | c | c | c c c c c}
    mask size & stride &  sampling & loss & CIDEr  & CIDEr-verb & CIDEr-arg & ROUGE-L & LEA
    \\
    \shline
    \hline
    \{1\} & 2s &  uniform & contrastive &  57.01 ± 2.21 & 63.80 ± 3.06 & 50.85 ± 1.77 & 41.69 ± 1.34 & \textbf{49.13 ± 1.24} % vbarg_denseslow_txe_txd_PretrainBackbone_K400_SLOW_8x8_R50_InfoNCE_train216k_epoch400_PretrainTXE_LVU_DENSESLOW_8x8_R50_InfoNCE_AGGTX_MaskPred_MaxMaskSize1
    \\
    \{1, 2\} & 2s &  uniform &  contrastive & 59.15 ± 2.01 & 66.79 ± 1.99 & 53.56 ± 1.61 & 42.16 ± 1.13 & 46.33 ± 0.94 % vbarg_denseslow_txe_txd_PretrainBackbone_K400_SLOW_8x8_R50_InfoNCE_train216k_epoch400_PretrainTXE_LVU_DENSESLOW_8x8_R50_InfoNCE_AGGTX_MaskPred_MaxMaskSize2
    \\
    \{1, 2, 3\} & 2s &  uniform & contrastive & \textbf{61.18 ± 1.48} &  \textbf{69.15 ± 1.57} & \textbf{55.00 ± 1.12} & \textbf{43.38 ± 0.87} & 47.81 ± 0.90 % vbarg_denseslow_txe_txd_PretrainBackbone_K400_SLOW_8x8_R50_InfoNCE_train216k_epoch400_PretrainTXE_LVU_DENSESLOW_8x8_R50_InfoNCE_AGGTX_MaskPred
    \\
    \{1, 2, 3, 4\} & 2s &  uniform & contrastive & 59.45 ± 1.31 &  64.71 ± 2.05 & 52.94 ± 1.55 & 43.25 ± 0.56 & 48.96 ± 0.61 % vbarg_denseslow_txe_txd_PretrainBackbone_K400_SLOW_8x8_R50_InfoNCE_train216k_epoch400_PretrainTXE_LVU_DENSESLOW_8x8_R50_InfoNCE_AGGTX_MaskPred_MaxMaskSize4
    \\
    \hline
    \{1, 2, 3\} & 1s &  uniform & contrastive & 60.15 ± 1.13 &  66.81 ± 1.64 & 53.29 ± 1.17 & 42.31 ± 0.68 & 47.66 ± 0.69 % vbarg_denseslow_txe_txd_PretrainBackbone_K400_SLOW_8x8_R50_InfoNCE_train216k_epoch400_PretrainTXE_LVU_DENSESLOW_8x8_R50_InfoNCE_AGGTX_MaskPred_ShotStride1
    \\
    \{1, 2, 3\} & 2s &  uniform & contrastive & \textbf{61.18 ± 1.48} &  \textbf{69.15 ± 1.57} & \textbf{55.00 ± 1.12} & \textbf{43.38 ± 0.87} & 47.81 ± 0.90 % vbarg_denseslow_txe_txd_PretrainBackbone_K400_SLOW_8x8_R50_InfoNCE_train216k_epoch400_PretrainTXE_LVU_DENSESLOW_8x8_R50_InfoNCE_AGGTX_MaskPred
    \\
    \{1, 2, 3\} & 3s &  uniform & contrastive & 60.22 ± 0.77 &  68.05 ± 0.67 &  53.48 ± 1.03 &  43.17 ± 0.49 &  \textbf{48.07 ± 0.69}  % vbarg_denseslow_txe_txd_PretrainBackbone_K400_SLOW_8x8_R50_InfoNCE_train216k_epoch400_PretrainTXE_LVU_DENSESLOW_8x8_R50_InfoNCE_AGGTX_MaskPred_ShotStride3
    \\
    \hline
    \{1, 2, 3\} & 2s &  uniform &  $L_2$ distance & 58.88 ± 1.13 &  66.43 ± 0.86 &  52.22 ± 1.01 &  43.05 ± 0.60 &  \textbf{48.55 ± 0.85} % vbarg_denseslow_txe_txd_PretrainBackbone_K400_SLOW_8x8_R50_InfoNCE_train216k_epoch400_PretrainTXE_LVU_DENSESLOW_8x8_R50_InfoNCE_AGGTX_MaskPred_L2Loss
    \\
    \{1, 2, 3\} & 2s &  max discrep. & contrastive & \textbf{61.65 ± 0.79} &  \textbf{68.44 ± 0.93} &  \textbf{55.06 ± 0.90} &  \textbf{43.44 ± 0.45} &  48.25 ± 0.54 % vbarg_denseslow_txe_txd_PretrainBackbone_K400_SLOW_8x8_R50_InfoNCE_train216k_epoch400_PretrainTXE_LVU_DENSESLOW_8x8_R50_InfoNCE_AGGTX_MaskPred_MultiMask_MaxDiscrepancy_WarmupEpoch10
  \end{tabular}
  \vspace{-2mm}
  \caption{\small  \it \textbf{Ablating mask prediction for TxE pretraining.} 
  The \textbf{top} section ablates the effectiveness of different sizes of the mask to apply (\eg, `\{1, 2\}' refers to sample mask sizes of 1 and 2). 
  The \textbf{middle} section ablates the impact of different token strides (\eg, `2s' refers to having two neighboring tokens as features computed from event segments that are 2s apart).
  The \textbf{bottom} section ablates alternative loss function and mask sampling strategy that adaptively select tokens to mask out. 
  }
  \label{table:vidsituablatemaskpred}
\end{table*}

\vspace{-5mm}
\paragraph{Impact of self-supervised pretraining.}
We present our results in Table~\ref{table:vidsitusrl}. The top section presents results from~\cite{sadhu-vidsitu2021}: {\tt \small GPT2} is a visual-blind language model baseline that only takes in verb classes as input; {\tt \small I3D+TxD} directly takes the I3D~\cite{carreira-i3d2017} video features as the input to the transformer decoder (TxD) without using the transformer encoder (TxE) to contextualize the features; and {\tt \small I3D+TxD+TxE} adds the TxE contextualizer to it. 
The second section of the table presents our results of different pretraining settings for the video backbone and TxE. Throughout, we use `N' to denote `not pretrained'.

% using a different backbone (Slow-D instead of I3) and a TxE architecture (unlike the TxE used in~\cite{sadhu-vidsitu2021}, we adopt residual multi-head attention modules and a learned positional embedding). We experiment with different backbones and TxE pretrainings (N = not pretrained). 

Several interesting observations arise from these results.
As reported in~\cite{sadhu-vidsitu2021}, the TxE contextualizer does not bring any performance gain over the simpler {\tt \small I3D+TxD} model (47.06 \vs 47.14 CIDEr) and our results using Slow-D show a similar trend (51.37 \vs 51.36). However, rather than drawing the conclusion that contextualization is not helpful in this case, we instead hypothesized that this is due to the lack of proper pretraining for TxE, and we could resolve this with our self-supervised pretraining using mask prediction on event features. Our results validate our intuition, as TxE pretraining significantly outperforms training TxE from scratch. The largest improvement comes from pretraining on the LVU dataset (which is 4.6x larger than VidSitu, therefore better performance), which improves CIDEr from 54.40 to 61.18.
% (Sec.~\ref{sec:lvu})
% This is thanks to 
This large gain comes from the fact that the mask prediction task essentially forces TxE to learn to contextualize input tokens (\ie, event features in this case) by propagating useful information among them. 
% \joe{maybe add something on why pretraining on LVU is better? Is it just dataset size?or how its constructed?}
While this has been demonstrated with large success in training language models like BERT, it has only been applied in vision in the form of predicting masked \emph{spatial regions} like patches~\cite{tan-vimpac2021}, objects boxes~\cite{wu-cvpr2021}, or joint vision-language pretraining that requires video-narration pairs~\cite{sun-videobert2019}. 
% \joe{should at least mention videobert too which may still be applied to small patches but I think is fairly well known work}
To the best of our knowledge, we are the first to show that this can be generalized to simple event-level video representations learned only using videos, and can lead to significant improvements over the state-of-the-art (61.18 \vs 47.14). 
% generalize well from language to visual transformer training. 
% \fy{add K400+VS \vs K400+K400 \vs VS+VS} 

In Sec.~\ref{sec:approach} we discussed the importance of separating pretraining for the backbone and the contextualizer to enable specialized training on the most suitable datasets. To quantify the importance of selecting the right dataset for each pretraining, we evaluate all possible permutations of using K400 and VidSitu (e.g., ``K400+VS'' refers to a CVRL backbone pretrained on K400, followed by a TxE pretrained using mask prediction on VidSitu). Among these, K400+VS achieves the highest performance (60.34), showing the importance of learning the backbone on the largest-scale dataset, but the TxE on an in-domain (i.e., movie) one. Interestingly, ``VS+VS'' performs the worst by a large margin, showing how the backbone does not need in-domain data to learn low-level video features and can generalize well to movies when pretrained appropriately.  
% \vspace{-1mm}

Finally, it is also promising to see that pretraining the backbone in a self-supervised fashion (CVRL-K400) generalizes better than fully supervised pretraining (Sup-K400): 54.40 \vs 51.36, which is likely due to the large gap between the supervised pretraining task and the downstream target task (action recognition \vs SRL in this case).  

\begin{table*}[h]
  \centering
  \small
  \begin{tabular}{l | c c c | c c}
    % \multicolumn{1}{c|}{} & \multicolumn{2}{c|}{VS-full} &
    % \\
    model & backbone pretrain & vb finetune & TxE pretrain & \hspace{0.3mm} Mean-Acc  & \hspace{0.3mm} Top1-Acc  
    \\
    \shline
    % \demph{Slow-only 8\x8 R50~\cite{sadhu-vidsitu2021}} & \demph{Sup-K400} & \hspace{1mm} \demph{29.05} & \hspace{1mm} \demph{29.05} 
    % \\
    \hline
    \demph{I3D~\cite{sadhu-vidsitu2021}} & \demph{Supervised} & \demph{\cmark} & \demph{-} & \hspace{1mm} \demph{34.13} & \hspace{1mm} \demph{39.91} 
    \\
    \hline
    Slow-D & Supervised & \xmark & - & \hspace{1mm} 33.03 ± 0.21 & \hspace{1mm} 41.90 ± 0.23 
    \\
    Slow-D & CVRL & \xmark & - & \hspace{1mm} 32.05 ± 0.24 & \hspace{1mm} 38.68 ± 0.69 % sfpret_onlyvid_evrel_PretrainBackbone_VidSitu_DENSESLOW_8x8_R50_Shots_InfoNCE
    \\
    Slow-D & MoDist & \xmark & - & \hspace{1mm} 33.29 ± 0.15 & \hspace{1mm} 40.52 ± 0.58 % sfpret_onlyvid_evrel_PretrainBackbone_VidSitu_VideoFlow_8x8_R50_Shots_InfoNCE_Epoch200
    \\
    \hline
    % \multicolumn{1}{c|}{-} & \cmark & \hspace{1mm} - & \hspace{1mm} -
    Slow-D & No & \cmark & - & \hspace{1mm} 30.66 ± 0.21 & \hspace{1mm} 41.29 ± 0.31 % sfpret_onlyvid_evrel_PretrainBackbone_kpret_denseslow_r50_16x4_Scratch
    \\
    Slow-D & Supervised & \cmark & - & \hspace{1mm} 34.00 ± 0.12  & \hspace{1mm} 40.65 ± 0.29 
    \\
    Slow-D & CVRL & \cmark & - & \hspace{1mm} 33.89 ± 0.17 & \hspace{1mm} 41.35 ± 0.24 % sfpret_onlyvid_evrel_PretrainBackbone_kpret_denseslow_r50_16x4_PretrainBackbone_VidSitu_DENSESLOW_8x8_R50_Shots_InfoNCE
    \\
    Slow-D & MoDist & \cmark & - & \hspace{1mm} 34.66 ± 0.18  & \hspace{1mm} 41.75 ± 0.47 % sfpret_onlyvid_evrel_PretrainBackbone_kpret_denseslow_r50_16x4_Pretrain_VidSitu_VideoFlow_8x8_R50_Shots_InfoNCE_Epoch200_epoch_6
    \\
    \hline
    Slow-D+TxE & CVRL & \cmark & No & \hspace{1mm} 25.00 ± 0.00  & \hspace{1mm} 39.42 ± 0.00 
    \\
    Slow-D+TxE & MoDist & \cmark & No & \hspace{1mm} 25.00 ± 0.00  & \hspace{1mm} 39.42 ± 0.00 % sfpret_onlyvid_evrel_PretrainBackbone_K400_SLOW_8x8_R50_InfoNCE_train216k_epoch400_PretrainTXE_VIDSITU_DENSESLOW_8x8_R50_InfoNCE_AGGTX_MaskPred_Backbone_K400_SLOW_8x8_R50_InfoNCE_train216k_epoch400
    \\
    Slow-D+TxE & CVRL & \cmark & MaskPred & \hspace{1mm} 34.71 ± 0.07  & \hspace{1mm} 41.16 ± 0.24 % sfpret_onlyvid_evrel_PretrainBackbone_CVRLK400VidSitu_SupVidSitu_PretrainTXE_VidSitu 
    \\
    Slow-D+TxE & MoDist & \cmark & MaskPred & \hspace{1mm} \textbf{35.32 ± 0.17}  & \hspace{1mm} 41.62 ± 0.43 % sfpret_onlyvid_evrel_freezeTxE_PretrainBackbone_MoDistK400_MoDistVidSitu_VbVidSitu_PretrainTXE_VidSitu_DENSESLOW_8x8_R50_InfoNCE_AGGTX_MaskPred
  \end{tabular}
  \vspace{-2mm} \caption{\small \it \textbf{Event relation prediction on VidSitu.} {\tt Supervised}: pretrained on K400 with class labels. {\tt CVRL}/{\tt MoDist}: pretrained with either CVRL~\cite{qian-cvrl2020} or MoDist~\cite{xiao-modist2021} first on K400 and then on VidSitu, as we found it helps bridging domain gaps to movies.  
  Methods in the \textbf{top} section are pretrained and then directly transferred to event relation prediction on VidSitu. 
  In the \textbf{middle} section, methods are further finetuned on VidSitu verb prediction task (vb finetune `\cmark'). 
  Finally, the \textbf{bottom} section shows the results of appending the transformer encoder (TxE) following the video backbone. 
  % Interestingly, TxE experienced training difficulties when initialized from scratch (it stuck in the local minimum with collapsed outputs), we contacted the authors of~\cite{sadhu-vidsitu2021} and they confirm they also observed similar behavior. However, when we then use mask prediction to pretrain TxE, it trains successfully and yield better accuracy. 
  % All models use a Slow-D 16\x4 R50 backbone. 
  For each experiment, we repeat 10 runs and report its mean and standard error. 
  }
  \label{table:vidsituevrel}
  \vspace{-5mm}
\end{table*}

\vspace{-5mm}
\paragraph{Ablating TxE pretraining.}
We now conduct ablation experiments to understand the impact of our design choices for TxE mask prediction pretraining (Table~\ref{table:vidsituablatemaskpred}). We experiment with the best model from Table~\ref{table:vidsitusrl}: [{\tt \small Slow-D+TxE+TxD, CVRL-K400, MaskPred-LVU}]. 
First, we study the impact of varying the size of the mask in Table~\ref{table:vidsituablatemaskpred} (top). Among all the tested options, uniformly sampling mask sizes from \{1, 2, 3\} achieves the best performance. Using a higher or lower value decreases performance considerably. This is understandable, as it would be too challenging to predict 4 masked out tokens out of a total of 5 tokens, meanwhile it would be too easy when too few tokens are masked out. 

Next, we study the impact of the stride size between two consecutive tokens (Table~\ref{table:vidsituablatemaskpred} middle). While all entries achieve competitive results, computing video features of events that are 2-seconds apart achieves the best balance.

Finally, we ablate on two other aspects of our mask prediction task in Table~\ref{table:vidsituablatemaskpred} (bottom): loss function and mask sampling strategy. 
First, we change the loss function from Eq.~\ref{eq:lmp} to a standard $L_2$ loss. This lowers the accuracy, likely due to $L_2$ being more sensitive to the representation collapse issue~\cite{grill-byol2020}.
Then, we compare the simple uniform sampling (used in all experiments) with a more sophisticated sampling strategy for mask positions: ``max discrepancy''. 
We believe that selecting good tokens to mask can help the model learn better. As an example, conceptually it would be more helpful for the model to learn to ``fill in'' a masked out event of a person showing painful expression from the preceding context event of another person punching with the fist, compared to masking out in the middle of a long shot of someone talking.   
% (\eg in Fig.~\ref{fig:approach} (right) it would be better to mask the rightmost token rather than the leftmost one). \joe{looking at the figure I don't find this obvious. is there another way to illustrate why you want to do this?}
%
Specifically, given a token $v_i$, discrepancy captures the difference in TxE output between $\hat{v_i}$ which is computed without any masking and $\hat{v}'_i$, which is computed by masking $v_i$. High discrepancy indicates that $v_i$ is an important token for TxE and masking it will push TxE to learn harder, using only the remaining tokens. Towards this, ``max discrepancy'' samples the token with the highest discrepancy. 
Surprisingly, this sampling strategy only achieve slightly better CIDEr score compared to the simple uniform sampling. We hypothesize that this is likely due to the nature of VidSitu, which contains short movie highlights, rather than more temporally coherent full movies and so the majority of tokens are already challenging if masked. %effective when we apply our method to full movies instead of movie highlights in LVU, as it's more important to sample ``interesting'' masks in that setting. 

\subsection{Event relation prediction} 
We now study the effectiveness of self-supervised pretraining on the even relation prediction task. This task is formulated as a 4-way classification problem between four relation types: ``A is enabled by B'', ``A is a reaction to B'', ``A causes B'', and ``A is unrelated to B''. Annotations are provided as  (A, B, relation) triplets. 
% for the relation between two events (2-seconds segments). \dvd{did not understand this last sentence}.
%
To predict the relation between events A and B, we concatenate features of both events either directly from the video backbone Slow-D (\ie, $v_A$ and $v_B$) or from the contextualizer TxE (\ie, $\hat{v}_A$ and $\hat{v}_B$). 
% The inputs to the classifier is the concatenation of the video feature computed from the backbone and the language representation of verbs, arguments and roles computed from a pretrained RoBERTa model~\cite{liu-arxiv2019}. 
We experiment with features from different models (Slow-D and Slow-D+TxE) and pretraining techniques in Table~\ref{table:vidsituevrel}. Following~\cite{sadhu-vidsitu2021}, we evaluate our results using the mean accuracy (averaged over relation types) and top-1 accuracy on the provided validation set. 
% \fy{Define these two metrics here}
% \dvd{we need to explain shortly what Macro-Averaged Accuracy is} \dvd{also, it's not clear from the table what vb finetune is. need to explain  that as well}

%
% (i) As recommended in ~\cite{sadhu-vidsitu2021}, we pretrained our backbone on K400 and further finetuned it on the verb prediction data of VidSitu (Sec.~\ref{sec:vp}) and, as expected, that helps boosting the performance (e.g., 34.66 \vs 33.29 with MoDist).
We present results for two settings: in the first one we directly transfer the pretrained features to event relation prediction (vb finetune `\xmark'). Whereas in the second setting, we take the models that are further finetuned on VidSitu verb prediction (`\cmark'), as done in~\cite{sadhu-vidsitu2021}. With both, self-supervised pretraining can be as effective as supervised pretraining, especially when using the motion-sensitive MoDist features (33.29\% \vs 33.03\%, and 34.66\% \vs 34.00\%). 
% vb finetune generally improves the event relation prediction accuracy, which shows the connection between these two tasks. 

Interestingly, we found that it does not work when naively adding the TxE and training it from scratch using randomly initialized weights, in which case the model does not train and reaches chance performance (Mean-Acc 25\%). This is likely why~\cite{sadhu-vidsitu2021} directly used the features from the video backbone and disregarded the outputs of TxE for this task. 
However, our results show that with proper pretraining using mask prediction, one can train Slow-D+TxE model for this task, and achieve better accuracy than using the backbone features alone: our mask prediction pretraining improves performance for both CVRL and MoDist, leading to state-of-the-art results of 35.32\% mean-acc compared to previous best result of 34.13\% reported in~\cite{sadhu-vidsitu2021}.

\begin{table}
  \centering
  \small
  \begin{tabular}{l | c | c c c}
    % \shline
    \multicolumn{1}{c|}{backbone} & pretrain & \hspace{-2mm} Acc@1  &  \hspace{-2mm} Acc@5 &  \hspace{-2mm} Recall@5
    \\
    \shline
    \demph{Slow~\cite{sadhu-vidsitu2021}} & \demph{Sup-K400} &  \demph{29.05} &  \demph{58.69} &  \demph{19.19} 
    \\
    \hline
    Slow-D & - &  31.69 &  68.64 &  5.68 
    \\
    Slow-D & Sup-K400 &  38.29 &  69.27 &  \textbf{18.70} 
    \\
    Slow-D & CVRL-K400 &  32.84 &  61.57 &  13.59 
    \\
    Slow-D & MoDist-K400 &  42.96 &  73.17 &  17.48 
    \\
    Slow-D & CVRL-K400+VS &  35.29 &  65.92 &  14.41 % kpret_denseslow_r50_16x4_PretrainBackbone_VidSitu_DENSESLOW_8x8_R50_Shots_InfoNCE
    \\
    Slow-D & MoDist-K400+VS &  \textbf{44.67} &  \textbf{74.38} &  18.40  % kpret_denseslow_r50_16x4_Pretrain_VidSitu_VideoFlow_8x8_R50_Shots_InfoNCE_Epoch200
  \end{tabular}
  \vspace{-3mm} \caption{\small \it \textbf{Verb prediction results on VidSitu.} 
  The first row shows the result applying the Slow-only network in~\cite{sadhu-vidsitu2021}.  
  % Throughout this work, we use a modified Slow-only network~\cite{qian-cvrl2020} (Slow-D for denser inputs) as the base to compare different pretraining methods in the bottom section. 
  We evaluate with top-1/5 accuracy as well as recall@5 metrics, following~\cite{sadhu-vidsitu2021}. For pretrain settings, we use the `method-dataset' notation. 
  % When comparing different self-supervised training methods, we found the motion-sensitive features produced by MoDist largely outperforms the RGB-only contrastive learning method CVRL. 
  % Further, when comparing self-supervised pretraining (CVRL and MoDist) to its supervised counterpart, we found SSL can work just as well. When pretrained on both K400 and VidSitu (no labels used), the MoDist representation outperforms supervised backbone with large margins for Acc@1/5 and on par for Recall@5. All models use a R3D-50 as its base.
  }
  \vspace{-3mm}
  \label{table:vidsituverb}
\end{table}

\begin{table*}[t]
	\setlength{\tabcolsep}{3pt}
	\centering
	\small
	\begin{tabular}{m{1cm} m{1cm} | c c c c c c c c c | c c}
		instance & scene & Relation($\uparrow$) &  Speaking($\uparrow$) & Scene($\uparrow$) & Director($\uparrow$)  & Writer($\uparrow$) & Year($\uparrow$) & Genre($\uparrow$) & Like($\downarrow$) & Views($\downarrow$) & \#top-1 & mean  \\
		\shline
		\hline
		OT~\cite{{wu-cvpr2021}}{\color{red}$^\dagger$} & \xmark & 50.00 & {\bf 34.57 }& 32.56 & 37.38 & 26.55 & 25.73 & 49.55 & 0.396 & 4.559  & 1/9  & 3.22 \\
		CVRL & \xmark & 50.95 & 32.86 & 32.56 & 37.76 & 27.26 & 25.31 & 48.17 & 0.444 & 4.600  & 0/9  & 3.89 \\
		MoDist & \xmark & 49.52 & 33.57 & 30.70 & 40.56 & 23.10 & 26.57 & 49.26 & 0.458 & 4.506  & 0/9  & 3.89 \\
		\hline
		\xmark & Sup & {\bf 52.38 }& 34.37 & 26.51 & 23.18 & 5.36 & 18.88 & 47.37 & 0.595 & {\bf 4.061 } & 2/9  & 4.44 \\
		\xmark & Ours & {\bf 52.38 }& 33.07 & 36.98 & {\bf 42.43 }& 23.93 & {\bf 35.24 }& 48.11 & 0.375 & 4.653  & 3/9 & 2.89 \\
		\hline
		OT~\cite{{wu-cvpr2021}} & Ours & 50.95 & 34.07 & {\bf 44.19 }& 40.19 & {\bf 31.43 }& 29.65 & {\bf 51.15 }& {\bf 0.353 }& 4.886  & {\bf 4/9 } & {\bf 2.33} \\
	\end{tabular} \vspace{-2mm}
	\caption{\small \it \textbf{LVU tasks.} 
  The first two columns show the pretraining setting we use for both the instance and scene representations (`\xmark' denotes `not used'). For each of the 9 tasks, we show the top-1 accuracy if it's a classification task, and mean squared error for regression tasks. In the last two columns, we show the number of tasks an approach gets the highest rank, and the mean rank of each approach across all tasks. {\color{red}$\dagger$}: note how the numbers of OT in the first row are lower than the ones reported in~\cite{wu-cvpr2021} because (a) we use a R50 instead of R101 backbone, (b) we use a Slow-D network for spatio-temporal features instead of the more expensive SlowFast network, and (c) we pretrain on 10k movies mentioned in their public repository instead of 30k movies used in their paper.
  }
  \vspace{-4mm}
	\label{table:lvu}
\end{table*}

\subsection{Verb prediction} 
\label{sec:vp}

Finally, we evaluate on the verb prediction task, which is the standard task of predicting action classes on short video segments.
 Each movie clip in the dataset is split into five 2-seconds event segments, each one annotated with a verb label. The dataset contains 1560 verb classes, like ``look'', ``talk'', ``walk'', ``run'', ``grab'', ``drive'', etc. 
We follow ~\cite{sadhu-vidsitu2021} and evaluate results using top-1 and top-5 accuracy (Acc@1/5) and top-5 recall (Rec@5) in Table~\ref{table:vidsituverb}.  
We observe the followings. 
(i) A self–supervised video backbone pretrained with MoDist outperforms a backbone pretrained with action labels on K400 (Acc@1: 42.96 \vs 38.29). This is surprising since some of the labels in this verb prediction task are the same as the action classes in K400. We argue this is due to the domain gap between YouTube-style action clips in K400 and movies in VidSitu, which self-supervised pretraining helps reducing. 
(ii) Between the two SSL methods, MoDist performs better than CVRL, suggesting its motion-sensitive features are well-suited for many verbs with strong motion like ``walk'' and ``run''.
(iii) Finally, when we extend to pretrain on both K400 and VidSitu (K400+VS), the performance further improves, as VidSitu helps to reduce the domain gap. While this is expected, one should note that it is only possible due to the self-supervised nature of the pretraining. 
To achieve a similar benefit using fully supervised pretraining, one would have to annotate videos in the new domain (e.g., genres for VidSitu movies), which is however expensive and non-scalable.

\section{Experiments: LVU Benchmark}\label{sec:lvu}
% \fy{Is it okay to submit ticket without this section?}
% \dvd{almost. just add a paragraph to introduce the dataset, its size and what it captures. so that the lawyers know what kind of dataset we are using}   

In this section, we demonstrate the effectiveness of our approach on the Long-form Video Understanding (LVU) dataset~\cite{wu-cvpr2021}. LVU is a large-scale dataset of 10k videos (typically 1-3 minutes long) with 9 diverse tasks, including user engagement ({\it YouTube like ratio}, {\it popularity}), movie meta data classification ({\it director}, {\it genre}, {\it writer}, {\it movie release year}) and content understanding classification ({\it relationship of actors in the scene}, {\it speaking style}, {\it scene}).
%Similar to VidSitu, LVU has sourced their video clips from MovieClips data~\cite{movieclips}. %As shown in~\cite{wu-cvpr2021}, it requires long-term temporal modeling to perform well on this benchmark, which makes it an ideal testbed for our model.

\begin{figure}[t]
    \centering
    % \hspace{-5mm}
    \includegraphics[width=0.45\textwidth]{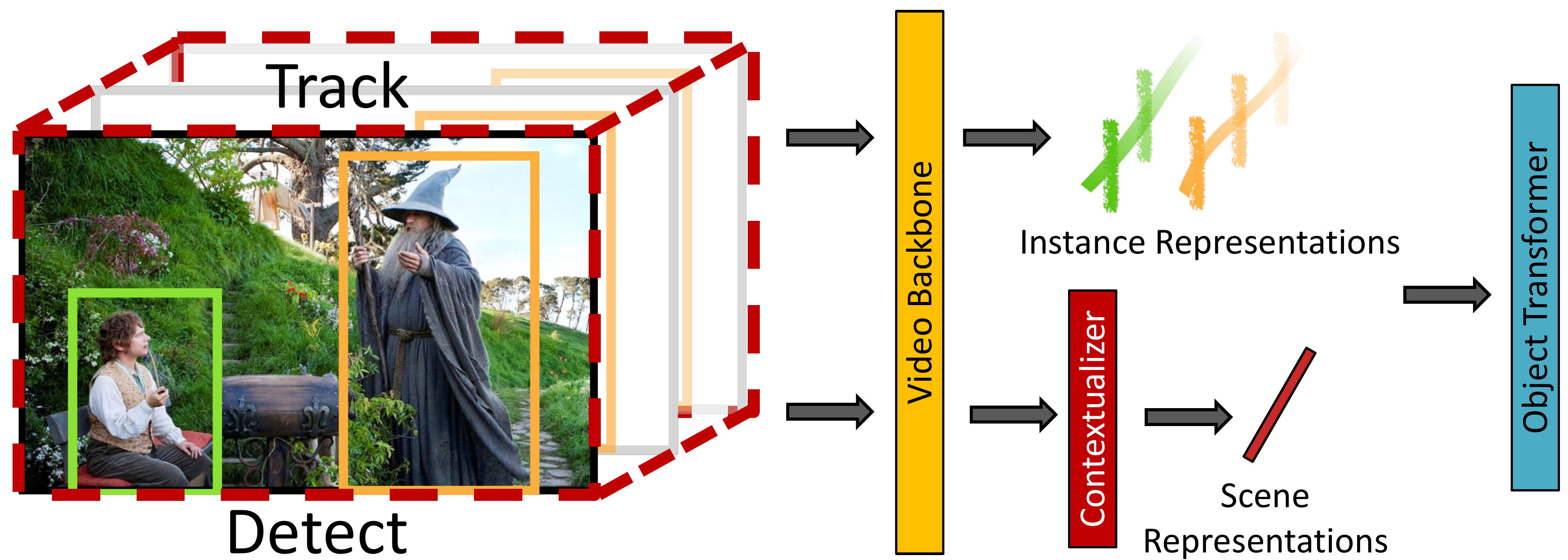} 
    \vspace{-2mm} \caption{\small  \it 
    %\textbf{LVU instance and scene representations}.
     \textbf{Object Transformer++}.
    Given the video and the detected/tracked objects, the top pathway feeds their cropped instance features into an Object Transformer to model their interactions~\cite{wu-cvpr2021}. 
    In addition to this, as shown in the bottom pathway, we propose to add a scene-level event representation, produced by our pretrained contextualizer (Sec.~\ref{sec:contextualizer}), 
    to model the context of the scene beyond the detected objects.   
    }
    \label{fig:lvu}
    \vspace{-4mm}
\end{figure}

% Their dataset is split into roughly 7k train, 15k validation and 15k test splits. 

% NOTES
% 1) introduce OT
% 2) talkes about what OT does -> SlowFast sup training
\vspace{-4mm}
\paragraph{Object Transformer++.} In~\cite{wu-cvpr2021}, the authors propose a long-term temporal model called Object Transformer (OT). It uses an object-centric design to represent each video as a set of spatio-temporal instances (i.e., tracklets of people and objects, Fig.~\ref{fig:lvu} top pathway) and a transformer-based architecture~\cite{vaswani2017attention} to model the synergies of the tracked instances in the video (blue rectangle). 
We argue that while such an object-centric design is useful for modeling long-term movie understanding, it is not sufficient. Reasoning only about the objects and the interaction among them can overlook the context of the scene, which is critical to understand movies (i.e., actors move out of the camera view, but the scene continues). Instead, we propose to enrich OT with a new scene representation. Specifically, we propose to use our self-supervised pretrained contextualizer (TxE, Fig.~\ref{fig:lvu} bottom pathway) to supplement the instance features. We denote the enhanced full method in Fig.~\ref{fig:lvu} as OT++.

\vspace{-4mm}
\paragraph{Results.} In Table~\ref{table:lvu}, we report the results  across all the 9 tasks proposed in the LVU dataset using the same parameters and experimentation protocol as~\cite{wu-cvpr2021}. We report the average performance over 5 runs. 
`instance' and `scene' are the two pathways of Fig.~\ref{fig:lvu}. The former denotes the object-centric features proposed by OT~\cite{wu-cvpr2021}, while the latter denotes the context features we propose as a mean to improve OT. 
% are Slow-D feature encoders that run on tracklets
`OT', `CVRL' and `MoDist' denote different features representing the instance tracklets (`instance' pathway). `OT' is first pretrained with full-supervision on K400 and then with instance masking on LVU, exactly as in \cite{wu-cvpr2021}. `CVRL' and `MoDist' instead are only pretrained with contrastive self-supervision on K400.
On the other hand, 'Ours' and 'Sup' instead encode whole frames (`scene' pathway). `Ours' denotes our self-supervised hierarchical pretraining, which consists of a Slow-D+TxE model with the backbone pretrained on K400 using CVRL, and TxE contextualizer pretrained with event mask prediction on LVU. While `Sup' is only a Slow-D backbone trained fully supervised on K400 without any contextualization (\ie, w/o the red block in Fig.~\ref{fig:lvu} bottom pathway). 
Finally, we use all these previously mentioned encoders to embed video clips and feed them to a final Object Transformer (blue box in Fig.~\ref{fig:lvu}) that is finetuned for the 9 LVU tasks.

In the first three rows we report the performance of using only the instance representation pathway, as is in~\cite{wu-cvpr2021}. 
The first row shows results for the OT baseline~\cite{wu-cvpr2021}. 
% In row 2-3, instead of using the supervised training on K400 and the instance masking task on LVU, we use backbone feature trained with only contrastive objectives (CVRL~\cite{qian-cvrl2020} and MoDist~\cite{xiao-modist2021}). 
As expected, features trained with only contrastive objectives (row 2-3) underperform~\cite{wu-cvpr2021}, which shows the effectiveness of OT for long term movie understanding. 
% \fy{I'm confused about this part, isn't the only difference the backbone feature here?}
% Furthermore, in row 4-6, we show the effectiveness of using different backbone features (\ie, no contextualization with transformers) with spatio-temporal pooling as our scene representation. In row 4, we use a supervised backbone trained on K400. Whereas in rows 5-6, we use backbone feature trained with CVRL~\cite{qian-cvrl2020} and MoDist~\cite{xiao-modist2021}.
%  (Sec.~\ref{sec:contextualizer})
However, when we use the contextualized features produced by pretraining with our event-level mask prediction task on LVU, even our scene representations w/o any instance features (Fig.~\ref{fig:lvu}, bottom pathway only) already outperform the much more complicated instance model OT (mean rank 2.89 \vs 3.22). When we combine this with the instance representation of OT, we obtain OT++, which achieves the best performance overall (mean rank 2.33), showing they are complementary. 
In addition, we also observe a significant improvement using our contextualized scene representations, compared to a scene representation directly pooled from the fully-supervised, but not contextualized,  backbone feature (row 5 \vs 4, mean rank 2.89 \vs 4.44), showing the importance of our pretrained contextualizer.   
Finally, we note that 3 tasks experience a performance drop when we combine instance and scene representations, likely because these tasks (\eg, predicting the year or director of a movie) do not rely on specific instance representations.

% \dvd{new point of Sup vs Ours scene pretraining. contexalization is useful}
% Furthermore, in row 4-6, we show the effectiveness of using different backbone features (\ie, no contextualization with transformers) with spatio-temporal pooling as our scene representation. In row 4, we use a supervised backbone trained on K400. Whereas in rows 5-6, we use backbone feature trained with CVRL~\cite{qian-cvrl2020} and MoDist~\cite{xiao-modist2021}. 

% However, when we use the contextualized features produced by pretraining with our event-level mask prediction task on LVU (Sec.~\ref{sec:contextualizer}), even our scene representations w/o any instance features (row 4) already outperform the much more complicated instance model OT (mean rank 2.67 \vs 3.00) \joe{I read 2.89 vs 3.22}. If combined with the instance representation from OT, the method achieves the overall best performance (mean rank 2.00), showing they are complementary. 
% \dvd{new point of Sup vs Ours scene pretraining. contexalization is useful}\joe{agree should point out that ours is better than sup in the text. I don't think you call out what sup is in the text either.}

\vspace{-1mm}
\section*{Conclusion}
We proposed a novel hierarchical self-supervised pretraining method tailored for movie understanding. 
Specifically, we proposed to separately pretrain each level of our hierarchical movie understanding model, so that they can become experts within the relevant domain (e.g., learn low-level appearance and motion patterns vs high-level contextualization). 
We demonstrated the effectiveness of our pretraining strategies on both VidSitu and LVU benchmarks, achieving state-of-the-art results.
% As we pretrain movie representations at multiple levels, we believe it can be widely useful for many more movie understanding tasks.
% beyond ones described in the paper. 
We hope these strategies will serve as a first baseline and encourage new research towards self-supervised learning for movies. 
%Of course, our work is not without limitation --- we demonstrated the effectiveness of our approach on trimmed movie contents (as VidSitu and LVU mostly consist of movie trailers and highlights), but can only hypothesize how will our method scale to untrimmed movies due to limited access to data. 
%That said, we believe our mask selection mechanism (`max discrep' in Table~\ref{table:vidsituablatemaskpred}) could potentially be more useful in that setting.   
Finally, for this research we used the following public codebases: PySlowFast~\cite{fan-pyslowfast2020}, VidSitu~\cite{sadhu-vidsiturepo2021} and LVU toolkits~\cite{wu-lvurepo2021}.

{\small
\bibliographystyle{ieee_fullname}
\bibliography{bibs}
}

\end{document}